\documentclass[runningheads]{llncs}
\usepackage[T1]{fontenc}
\usepackage{graphicx}
\usepackage{amsmath}
\usepackage{amssymb}
\usepackage{booktabs}
\usepackage{blindtext}
\usepackage{multirow}
\usepackage{placeins}
\usepackage{makecell}
\usepackage{stfloats}
\usepackage{letltxmacro,xparse}
\usepackage[table]{xcolor}
\usepackage{array}
\usepackage{hyperref}

\usepackage[capitalize]{cleveref}
\crefname{section}{Sec.}{Secs.}
\Crefname{section}{Section}{Sections}
\Crefname{table}{Table}{Tables}
\crefname{table}{Tab.}{Tabs.}

\usepackage{color}

\newif\ifreview
\reviewfalse

\ifreview
	\usepackage{lineno}

	\linenumbers
\fi

\newcommand{\encoder}{\mathbf{E}}
\newcommand{\decoder}{\mathbf{D}}
\newcommand{\downstream}{\mathbf{F}}

\begin{document}


\def\SubNumber{25}

\def\GCPRTrack{Special Track: Photogrammetry and remote sensing}

\title{SenPa-MAE: Sensor Parameter Aware Masked Autoencoder for Multi-Satellite Self-Supervised Pretraining}

\ifreview
\else
	\titlerunning{SenPa-MAE: Sensor Parameter Aware Masked Autoencoder}

	\author{Jonathan Prexl\orcidID{0009-0006-2560-5817} \and Michael Schmitt\orcidID{0000-0002-0575-2362}}
	
	\authorrunning{J. Prexl and M. Schmitt}
	
	\institute{Department of Aerospace Engineering, University of the Bundeswehr Munich, Munich, Germany
    \email{\{jonathan.prexl, michael.schmitt\}@unibw.de}}
\fi

\maketitle              

\begin{abstract}
This paper introduces \textit{SenPa-MAE}, a transformer architecture 
that encodes the sensor parameters of an observed multispectral signal into the image embeddings. \textit{SenPa-MAE} can be pre-trained on imagery of different satellites with non-matching spectral or geometrical sensor characteristics.
To incorporate sensor parameters, we propose a versatile sensor parameter encoding module as well as a data augmentation strategy for the diversification of the pre-training dataset.
This enables the model to effectively differentiate between various sensors and gain an understanding of sensor parameters and the correlation to the observed signal. 
Given the rising number of Earth observation satellite missions and the diversity in their sensor specifications, our approach paves the way towards a sensor-independent Earth observation foundation model. This opens up possibilities such as cross-sensor training and sensor-independent inference.
The code for model training as well as the model weights are publicly available under
\url{https://github.com/JonathanPrexl/SenPa-MAE}.

\keywords{Earth Observation \and Multi-Spectral \and Deep Learning \and Self-Supervised \and Sensor Parameter Encoding}
\end{abstract}

\section{Introduction}
\label{sec:intro}
Optical systems that map the three-dimensional world onto a two-dimensional image plane are characterised by a set of sensor-specific parameters that determine the final appearance of the image. In many applications of classical computer vision, it is desirable to be agnostic to the specific camera parameters, since - during inference - a model should work on a broad range of sensors (e.g. pictures taken with different cameras).
Often, this is even specifically enforced by image augmentations during the training phase which makes models invariant to the applied augmentations for a given input image. Nevertheless, in some imaging domains, the specific camera parameters do carry crucial information about the observed scene and should not be neglected.
In those cases, the image data should be interpreted as physical measurements of a process with the results being encoded in a tensor $\boldsymbol{x}$ of relevant physically interpretable parameters (emission or reflectance values measured in a specific section of the electromagnetic spectrum), rather than the vague concept \textit{image}.
In those cases, it is desirable that machine learning models are aware of the specific sensor parameters, especially when multiple different sensors are used, or when the measurements are subject to a subsequent data fusion or comparison step.

A prime example, where specific sensor parameters are essential for the interpretation of the data at hand is the field of satellite-based Earth observation (EO). Here, since the looking angle can be viewed as constant (for moderate-resolution imagery) the measurement is mainly determined by two sensor-specific properties, which specify the spatial and spectral resolution of the underlying imagery.
The point spread function (\textit{PSF}) of an optical system describes the sensor response at the imaging plane when illuminated by a point-shaped light source. Since geometrical sensor resolution is defined by the sensor's capability to separate two point sources with distance $d$ between each other, the \textit{PSF} directly determines the resolution limit and therefore the physically feasible pixel spacing of the sensor.
The domain-specific EO term for the resulting pixel-spacing in the context of satellite imagery is the so-called \textit{ground-sampling distance} (GSD), whose values we will in the following denote as $\sigma_c$, with $c$ being the channel index.
Secondly, optical systems are in many cases designed to measure multiple different portions of the electromagnetic spectrum, represented by multiple channels (or bands). The number of channels for an optical system can vary and - in the case of optical moderate resolution EO - is often $\mathcal{O}(10)$. Here next to the visible RGB channels, further channels are designed to measure exact biochemical processes on the Earth's surface often outside of the visible spectrum, mostly in the near and short wave infrared regime \cite{drusch2012sentinel,pahlevan2022simultaneous}.
Every channel $c$ is defined by a so-called spectral response function which we will denote as $\boldsymbol{\lambda}_c$. The response function of a band describes how sensitive the corresponding band is to the incoming light of a given wavelength, determined by the absorption and reflectance characteristics of the optical components as well as the quantum efficiency of the sensor. 
Therefore, an optical system onboard a satellite can be characterized by the set of all corresponding $\{\boldsymbol{\lambda}_c, \sigma_c\}$ as well as the revisit time which is determined by the satellite orbit and optical swath. EO missions are always subject to an engineering tradeoff between these three quantities and cannot be optimized for all three while keeping a high signal-to-noise ratio. Therefore, missions are usually tailored towards specific demands and the analysis and fusion of multiple different data sources bring significant benefits \cite{claverie2018harmonized}.
\begin{figure}
    \centering
    \includegraphics[width=1\textwidth]{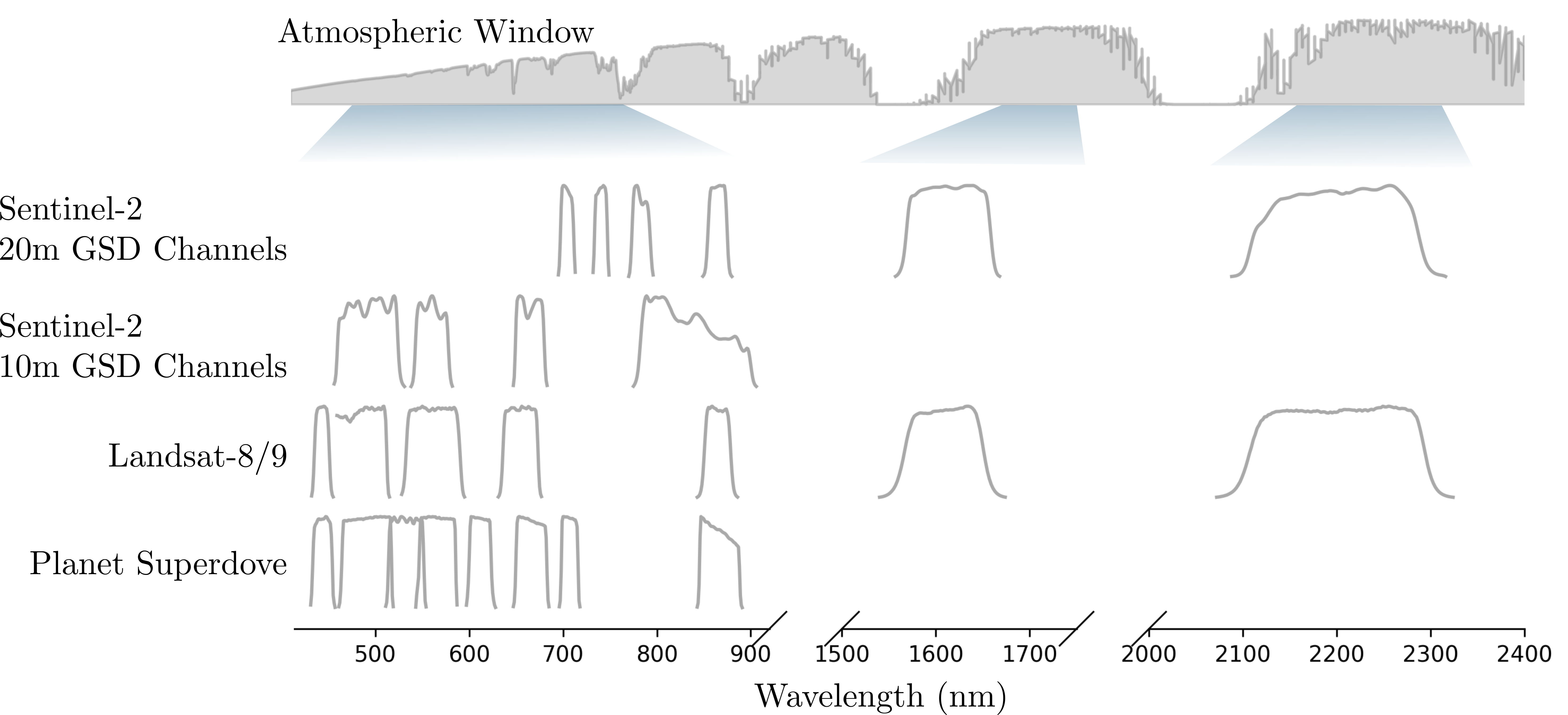}
    \caption{The spectral response functions $\boldsymbol{\lambda}_c$ for the Sensors \textit{Landsat}, \textit{Sentinel-2} and \textit{Planet SuperDove} as a function of the wavelength. Each $\boldsymbol{\lambda}_c$ has values ranges in $[0,1]$, and shifts along the $y$ axis are purely to distinguish different sensors and different sets of $\sigma_c$ within the set of channels. The atmospheric transmittance spectrum (ATS) of the Earth, crucial when designing channels, is provided as a reference. Shaded blue boxes below the ATS indicate the position of the channels relative to the ATS.}
    \label{fig:responsefunctions}
\end{figure}

The amount of publicly available Earth observation imagery provided by state agencies and corporate entities has grown rapidly over the last decade \cite{chi2016big,SchmittData}. 
Some of the most commonly utilized missions are \textit{Landsat} (LS) and \textit{Sentinel-2} (S2) provided by \textit{NASA} and \textit{ESA}, respectively, as well as data from the constellation of the privately founded company \textit{Planet} --e.g. acquired by their newest sensor \textit{SuperDove} (SD).
 To give the reader an intuition about the differences in sensor design, \cref{fig:responsefunctions} compares the spectral response functions $\boldsymbol{\lambda}_c$ for all three above-mentioned sensors.
The data collected by these three sensors serves as the cornerstone for numerous socially impactful research projects and applications, such as disaster mapping \cite{amatya2022rainfall}, agriculture applications \cite{segarra2020remote},  wildfire monitoring \cite{filipponi2019exploitation}, and environmental studies \cite{yuan2020deep}.

To this day, machine learning models for EO data applications are usually trained to work on one specific sensor \cite{SchmittData}. An approach to generalize a model over a set of sensors holds a large potential, since multiple training data sets, built on top of different sensors, could be utilized. Further, it is desirable to perform inference in a sensor-independent manner, especially in time-critical scenarios, e.g. disaster-extent mapping where the revisit time and the availability of data play a crucial role.
Still, merging data and labels of multiple sensors is a non-trivial task, due to the differences in spectral characteristics, resolution and acquisition date. 
Naively training across multiple sensors without accounting for their distinct differences would result in the introduction of invariances with respect to the sensor-specific parameters.
Consequently, models trained in this fashion would disregard the nuances among sensors that are meticulously designed to provide optimized data for specific applications.

In this work, we present a flexible way of incorporating sensor-specific parameters and performing pre-training on a set of various sensors.
Models pre-trained with our approach hold the potential to generate sensor-independent EO foundation models that can be fine-tuned towards sensor-independent inference on downstream applications.
We perform pre-training of a ViT-based transformer architecture with the masked autoencoding strategy while introducing a flexible way to inject sensor parameters. We enforce the relevance of these sensor parameters in the pre-training phase by adapting the masking function and the data augmentation module in a way suitable to fit the peculiarities of multi-spectral imagery.
The effects of our proposed method with respect to the application of predicting land cover in various multi-sensor settings are investigated in the experiment section of this manuscript.  

\section{Related Work}
\label{sec:relatedwork}

This section provides a comprehensive overview of transformer architecture, focusing on their application to computer vision and remote sensing. We delve into effective pretraining strategies and discuss adaptations required for remote sensing tasks.

\noindent\textbf{Transformer} The transformer architecture \cite{vaswani2017attention} is known for revolutionizing influence on the field of natural language processing (NLP) \cite{kenton2019bert}.
This is mainly due to the flexibility given by the sequence-to-sequence character of the architecture while featuring high-performance properties due to better parallelization capabilities and the capturing of long-range dependencies. Those properties lead to remarkable improvements in NLP tasks but their application has also been shown to extend to computer vision. 
After initial tries to combine the attention mechanism with - up to that point state of the art - CNN architectures \cite{bello2019attention,carion2020end,tian2020attention},
the introduciton of \textit{Vision-Transformer} (\textit{ViT})  \cite{dosovitskiy2020image} represented the first comparably performing pure transfomer approch.
Here, the strategy builds on maintaining proximity to the NLP application by dividing the image into a sequence of patches where the corresponding patch embeddings (tokens) can be processed by transformer layers without additional alterations. To this day, ViT matches and exceeds the benchmark for many image classifications downstream tasks. 

\noindent\textbf{Pre-Training of Transformers} A very promising pre-training strategy for ViT type architectures was introduced by the\textit{Masked Autoencoder} (MAE) \cite{he2022masked}.
Here, an encoder network is used to process a randomly chosen subset of $25\, \%$ of the tokens and subsequently, a decoder network is asked to restore the original image, given the embeddings generated by the encoder. This seemingly difficult training objective represents a challenging self-supervision pretext task which requires image understanding beyond the low-level image statistics, hence producing descriptive features. MAE - up to this day - constitute the most popular pre-training strategy and the authors could demonstrate state-of-the-art performance on various image-classification tasks and show the importance of the masking strategy \cite{he2022masked}.
Authors of \textit{MultiMAE} \cite{bachmann2022multimae} extended this concept to a cross-modal (RGB image, depth-map, semantic-map) learning setup. Here, task specific decoders can reconstruct the corresponding cross-modal information from the decoded image features, showing the spatial- and cross-modal predictive capabilities of the MAE pre-training paradigm.

\noindent\textbf{Domain adaption to Remote Sensing}
Satellite-based Earth-Observaion (EO) is a domain with a nearly unlimited amount of unlabeled data. Still, the amount of datasets with high-quality labels is very limited. Hence, pre-training networks on unlabeled data and finetuning for solving specific downstream tasks hold huge potential. 
One of the first works that study pre-trained transformers for the field of EO is \textit{SatVit} \cite{fuller2022satvit} which is pre-trained on channel-wise concatenated optical and synthetic-aperture radar (SAR) images. The author shows the gain in performance for peatland and landcover prediction when finetuning from a pre-trained network.
In subsequent works like \textit{SatMAE} \cite{cong2022satmae} the authors restrict the input purely to the optical domain but emphasize the domain-specific requirements of multi-spectral EO data. Detailed encoding strategies for unequally distributed time series (a common occurrence in RS tasks) as well as the introduction of channel groups (separate tokens for sets of similar channels) are introduced.
In \textit{ScaleMAE} \cite{reed2023scale} the Authors work with imagery of different resolutions and encode the land area covered by the image patches into the embeddings while reconstructing multiscale representations on various resolution levels.

During the final preparation of this manuscript, researchers in \cite{xiong2024neural} introduced an alternative method for processing multispectral and multimodal input images using a vision transformer. This approach utilizes a hyper-network to dynamically generate the weights of the patch embedding layer according to channel wavelengths. This allows for flexible handling of varying channel counts or different modalities while maintaining a consistent sequence length within the transformer.

While the aforementioned methods consider some specific characteristics of EO data, we aim to enhance generalizability by directly incorporating the relevant camera parameters into the derived image embeddings via a simple and generalizable strategy. With the increasing diversity of EO missions, this approach paves the way for a more universal pretraining paradigm for EO foundation models.

\section{Method}
\label{sec:method}

The objective of the \textit{SenPa-MAE} architecture proposed here is to generalize the prediction step that usually processes an image tensor $\mathbf{x}$ to the more general setup of processing $\mathbf{x}$ together with the corresponding sensor parameters for a complete description of the observed signal. Within this study, the sensor specifications are described by the set of corresponding spectral response functions $\{ \boldsymbol{\lambda}_1,\hdots,\boldsymbol{\lambda}_C \}$  and the GSD values $\{ \sigma_1,\hdots,\sigma_C \}$ per channel. In the following, we will denote the corresponding set of sensor parameters for a given multispectral signal $\mathbf{x}$ as $\{ \boldsymbol{\lambda}_c, \sigma_c \}$.
This generalization step opens up the possibility of processing data from various sensors with different specifications within one model.
We will use this to conduct pertaining of the network over various different sensors and later test the capability of the model to work with data from different sensors when finetuned to a downstream application. 
Generally, the pre-training is based upon the MAE training paradigm \cite{he2022masked} and can be conducted in a self-supervised fashion over a dataset containing images from multiple sensors without needing to have matching observations (i.e. with the same acquisition date and location), as these are hard to get due to the different revisit time schedules of the satellite missions.

\begin{figure}[t!]
    \centering
    \includegraphics[width=1\textwidth]{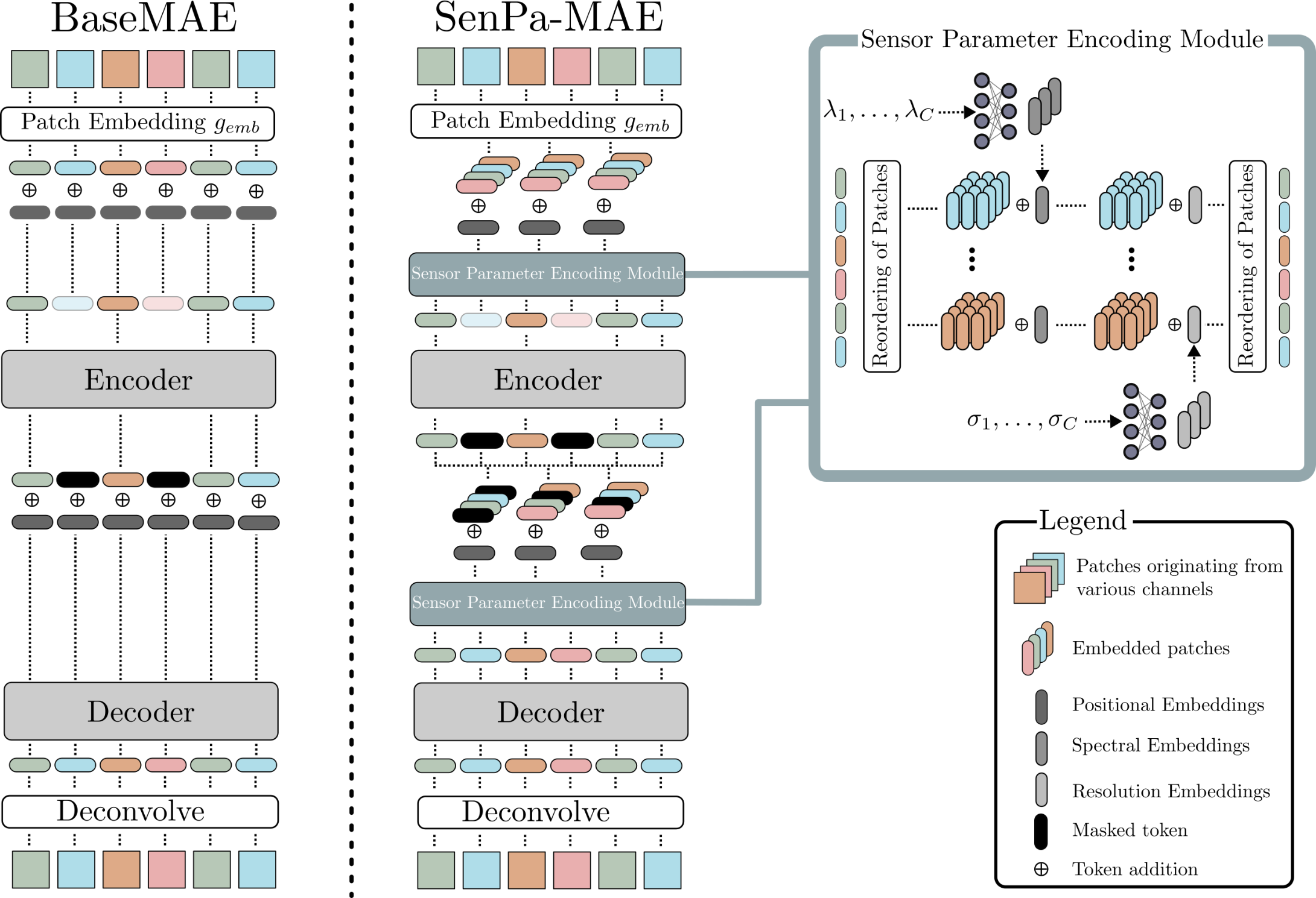}
    \caption{The proposed \textit{SenPa-MAE} architecture as well as the baseline model \textit{BaseMAE}. After patch embedding, the tokens $\mathbf{t}_i$ undergo a three-step encoding procedure where information about patch position, spectral response function, as well as ground sampling distance, gets added. After the encoding, all tokens get processed in an MAE-like encoder decoder setup. An analogue encoding procedure can be applied before the decoding step (compare \cref{sec:CPI}). The \textit{BaseMAE} setup differs from \textit{SenPa-MAE} by purely taking the positional encoding into account and therefore, neglecting the injection of the sensor parameters.}
    \label{fig:mae_ra_architecure}
\end{figure}

In the following subsections the two major technical contributions of \textit{SenPa-MAE}, namely the sensor-parameter encoding module and a novel data augmentation method will be introduced in detail.  
To guide the reader, \cref{fig:mae_ra_architecure} gives a graphical overview of the two proposed architectural modifications. 

\subsection{Sensor Parameter Encoding}
\label{sec:CPI}

Injecting multiple non-equally-shaped types of information (i.e. image and corresponding sensor parameter) into a neural network is a non-trivial task. We build upon the ViT architecture \cite{dosovitskiy2020image} and the associated positional encoding strategies to inject the 3D multi-spectral data tensor $\mathbf{x}$ of size $H\times W$ with $C$ channels and the set of corresponding response functions $\{ \boldsymbol{\lambda}_1,\hdots,\boldsymbol{\lambda}_C \}$  and GSD values $\{ \sigma_1,\hdots,\sigma_C \}$ into the model.  
The response function information is gathered from the satellite providers and resampled to $300\,$nm - $2600\,$nm wavelength interval, with $1\,$nm step width.
Therefore each channel has a corresponding scalar GSD value $\sigma_c \in \mathbb{R}^{1}$ and as well as a response function vector $\boldsymbol{\lambda}_c \in \mathbb{R}^{2300}$.

In our framework, when the parameter encoding module is active, the model input takes the form $(\mathbf{x},\{ \boldsymbol{\lambda}_c, \sigma_c \})$. This means that each multispectral image tensor $\mathbf{x}$ is supplemented with the corresponding sensor parameters. This approach is applied both during the model pre-training and for subsequent fine-tuning on downstream tasks.

\noindent\textbf{Patch Encoding and Masking} Unlike the conventional ViT approach, where an image is segmented into patches of size $P^2$, each patch having a position $(i,j)$, and all channel information for a patch is encoded into a single token, our method generates a token for each patch and channel. This results in $N=C \times HW/P^2$ tokens $\mathbf{t}_i$ of dimension $D_{emb}$.
This design decision is motivated by two facts: In contrast to regular RGB images, for multispectral signals much of the desired information content of multispectral Earth observation data lies in the relative reflectance statistic of individual bands. Therefore, providing single-channel tokens enhances the flexibility of applying the attention mechanism across channels.
Furthermore, this presents the opportunity to mask a significant portion of the tokens for the MAE masking step without creating extensive regions in the image devoid of geometric information. This is especially crucial since optical satellite images typically capture objects at a relatively small scale (a few pixels). In contrast to that, in regular RGB imagery, objects often occupy the entire image, making reconstruction of heavily masked images with extensive contiguous masked areas a viable task.

\noindent\textbf{Multispectral-ViT Encoding} Given all tokens $\mathbf{t}_i$ with $i \in (0,\hdots, N)$ we first apply a learnable positional encoding vector $\boldsymbol{\omega}^{pos}\in \mathbb{R}^{N/C \times D_{emb}}$ to each token originating from one position in the multispectral tensor $(i,j)$.
This only encodes the position in the image since tokens originating from different channels still get added with the same positional encoding. Subsequently, to the positional encoding we apply the sensor parameter encoding module. Here, we use the $C$ corresponding spectral response function $\{ \boldsymbol{\lambda}_1,\hdots,\boldsymbol{\lambda}_C \}$ and GSD values $\{ \sigma_1,\hdots,\sigma_C \}$ and encode them into the embedding dimension by applying two separate five-layer MLP networks $g^{SRF}$ and $g^{GSD}$
\begin{equation}
\boldsymbol{\omega}^{SRF}_{c} = g^{SRF}(\boldsymbol{\lambda}_c) 
\end{equation}
\begin{equation}
\boldsymbol{\omega}^{GSD}_{c} = g^{GSD}(\sigma_c)
\end{equation}
which leads to the spectral and resolution encoding vectors  $\boldsymbol{\omega}^{SRF}_{c}\in \mathbb{R}^{D_{emb}}$ and $\boldsymbol{\omega}^{GSD}_{c}\in \mathbb{R}^{D_{emb}}$. We reorder the tokens $\mathbf{t}_i$ and add $\boldsymbol{\omega}^{SRF}_{c}$ and $\boldsymbol{\omega}^{GSD}_{c}$ to all tokens that originate from the corresponding channel $c$. 

With the set of tokens with positional, spectral and resolution encoding applied, the image features $\mathbf{f}_i$ get created by applying an encoder network $\mathbf{f}_i = \encoder(\mathbf{t}_i)$  to a random subset of tokens determined by the masking ratio $r_{mask}$.
Analogue to the MAE strategy we fill up the endoced tokens by learnable mask-tokens $\mathbf{t}_{mask}$ to restore the original sequence length. 

We test two different modes of encoding strategies for the decoder step. In both scenarios, analogue to the encoding step the features $\mathbf{f}_i$ get summed up with a positional encoding vector $\boldsymbol{\omega}^{pos} \in \mathbb{R}^{N/C \times D_{emb}}$ depending on the position of origin in the image $(i,j)$. We then examine the two scenarios: one where the sensor parameter encoding module is applied a second time and one where it is not.
This follows from the intuitive idea that not encoding the sensor parameters in the decoding step forces a model to implicitly encode the information into the features. In contrast to that, applying a second sensor parameter encoding after the decoding stage would provide the decoder $\decoder$ with the necessary information for the reconstruction task and therefore the information is not necessarily encoded in the features $\mathbf{f}_i$.
After the second encoding step, the features get processed by the decoder network $\decoder$ and the decoded features get transformed into the original patch dimension by a trainable linear layer.

\noindent\textbf{Baseline} To ensure a fair evaluation of the above-mentioned contribution we propose a standard MAE-like baseline \textit{BaseMAE}. Here, the parameter encoding module gets neither applied before the decoding nor the encoding network. This makes an adaption of the positional encoding necessary since otherwise, the position of the tokens is not uniquely determined. Therefore we extend the positional encoding to a three-dimensional version $\boldsymbol{\omega}^{pos}\in \mathbb{R}^{N \times D_{emb}}$. 
This encodes the position in the image and the channel index in the image tensor $\mathbf{x}$ without taking the sensor parameters into account. If no augmentation to the channel order is applied this leads to a sensor-specific input order and hinders the flexibility of applying it to multiple inputs. As later discussed in \cref{sec:ssa} we therefore work with an arbitrary order of channels. 

\subsection{Spectral Superposition Augmentation}
\label{sec:ssa}

Diversity and amount of data during the training of neural network architectures are known to enhance the performance significantly.
Still -- as described above -- producing models that are invariant to image manipulations like changing brightness or contrast is not an option when pursuing the objective of generating sensor parameter-aware architectures. 

\noindent\textbf{Spectral Augmentation} We propose a data augmentation strategy which we further call \textit{Spectral Superposition Augmentation} (SSA). Here we build synthetic multispectral signals $\mathbf{\Tilde{x}}$ by the superposition of two randomly selected channels within the set of channels from one sensor:
\begin{equation}
    \mathbf{\Tilde{x}} = \alpha_1 \mathbf{x}_m + \alpha_2 \mathbf{x}_n
\end{equation}
where the corresponding spectral response function for $\mathbf{\Tilde{x}}$
adds up to:
\begin{equation}
    \Tilde{\boldsymbol{\lambda}} = \alpha_1 \boldsymbol{\lambda}_m + \alpha_2 \boldsymbol{\lambda}_n \text{ .}
\end{equation}
This augmentation strategy increases the diversity of examples during training, can easily be extended to combine more than two channels,  and requires a detailed understanding of $\boldsymbol{\lambda}$ to reconstruct images. We apply this augmentation step to a random number of channels in the input image determined by $p_{mix}$.

\noindent\textbf{Resolution Augmentation} To further enhance the diversity of data seen during training we additionally perform a cubic downsampling operation on $\mathbf{x}_c$ and $\Tilde{\mathbf{x}}_c$.
Starting from the original GSD of the sensor $\sigma_{org}$ we randomly select a target GSD from the set $\{ \sigma | \sigma > \sigma_{org}, \sigma \in (5,10,15,20,30) \}$ for a random subset of images (downsample probability $p_{down}$ per channel) in the batch.
Prior to the downsampling operation, we apply Gaussian blurring in order to realistically mimic a sensor with lower GSD values. 

\section{Technical Details}
\label{sec:TD}
In this section, we describe the data used for the pre-training of the model and the subsequent downstream task evaluation, as well as the technical implementation details for those two steps. 

\noindent\textbf{Pre-training Dataset} The pre-training dataset consists of all $10\,$m and $20\,$m GSD bands of \textit{Sentinel-2} (S2), \textit{Planet-Superdove} (SD) and \textit{Landsat-8/9} (LS) imagery globally sampeld from 29 locations. All used bands are displayed in \cref{fig:responsefunctions}. Those three satellite missions represent the most frequently used sources for optical Earth observation data and differ in their number of channels, the corresponding response functions and resolutions. While access to S2 and LS imagery is free, the \textit{Planet-Superdove} can be acquired through \textit{Planets} free research plan.
All scene IDs of the used imagery can be taken from the project repository. 
From the 29 locations we overall sample 50k patches of size $256\,\text{px} \times 256\,\text{px}$ for the S2 and SD images, as well as 25k further patches from LS. We will apply a geographical split and hold out 3 of the 29 locations for validation purposes, ending up with $~116$k images from three sensors used for training. 

The GSD values for the above-described imagery span the values $\sim 3.5\,$m for SD to $10\,$m and $20\,$m for S2 and $30\,$m for LS.
Since we want to encode the sensor resolution into the model - independent of the covered area - we resample (cubic) all images to a $5\,$m pixel-spacing grid, therefore keeping the ground extent of the input imagery constant.

\noindent\textbf{Finetuning Land Cover Dataset} Similar to the above-described process we gather spatial and temporal overlapping S2 and SD images around nine locations in the United States. After a regional split, five locations are used for training with $\sim 42$k patches of size $256\,\text{px} \times 256\,\text{px}$, per sensor. \textit{ESA-World Cover V200} \cite{zanaga2022esa} data for the patches is collected and, after removing non-occurring classes, an eight-class classification scheme is chosen. 
Analogue to the pre-training all data is resampled to a $5\,$m pixel-spacing.
The distribution of the locations for the pre-training as well as for the fine-tuning dataset can be taken from the \textcolor{black}{supplemantary materials}.

\noindent\textbf{Pre-Training Model Settings} We train \textit{SenPa-MAE} and \textit{BaseMAE} to reconstruct the unmasked input with a masked mean absolute error loss. All models have a fixed size of $12$ layers with $12$ heads for the encoder and $3$ layers with $12$ heads for the decoder, respectively, which corresponds to the ViT-Base setup in \cite{dosovitskiy2020image}. 
Since the satellites in our study have varying numbers of channels, determining the input length for the model is not straightforward. Therefore, we decided to train all models using four randomly selected channels. This maintains a consistent number of input channels and ensures that the model is exposed to all available information from the different channels, preserving its generalizability. 
We sample randomly $144 \, \text{px} \times 144 \, \text{px}$ input images from the raw $256 \, \text{px} \times 256 \, \text{px}$ data and generate a $9\times 9$ patch pattern per channel. All patches are transformed into the latent token space with size $D_{emb}=768$.
The masking ratio $r_{mask}$ is set constant to $66\%$ for all runs. If the data augmentation module (SSA) is active the probabilities for channel mixing $p_{mix}$ and downsampling $p_{down}$ are set to $25\%$, where we randomly choose two or three channels for the spectral superposition.
Training takes place over $400$ epochs with $3$ warmup-epochs, a cosign decay learning rate scheduler (initial learning rate of $1 \times 10^{-4}$), a \textit{Adam optimizer}, a weight decay of 0.05 and processed with a batch size of $128$.

\begin{figure}
    \centering
    \includegraphics[width=0.6\textwidth]{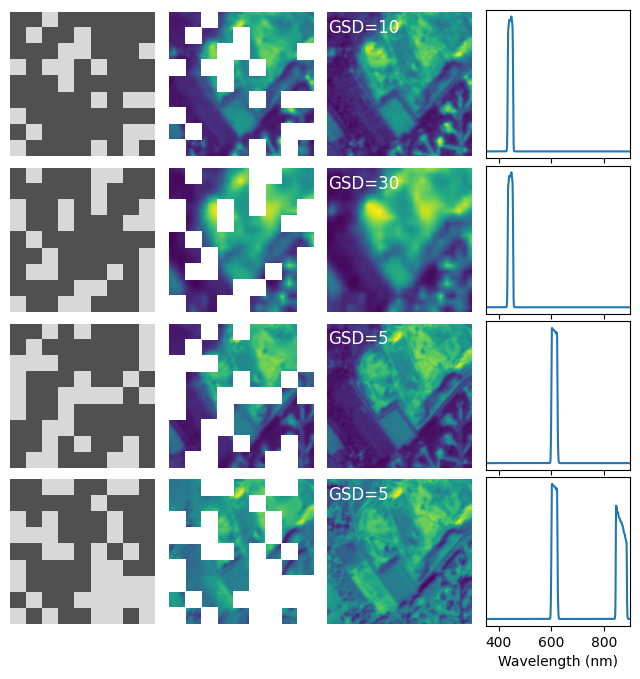}
    \caption{The reconstruction result for a masked four-channel input. From left to right: Channel-specific mask, reconstructed signal for the masked areas, ground truth image with the GSD of the channel, as well as the spectral response function.}
    \label{fig:reconstruction}
\end{figure}

\noindent\textbf{Reconstruction Quality}
Given the above-described data augmentation strategy, masking strategy and model settings, exemplary results of the reconstruction task for active data augmentation (channel superposition and active downsampling) are shown in \cref{fig:reconstruction}.
Here, it can be seen that the model successfully reconstructs input while successfully adjusting to the different sensor parameters.
More visual examples along different training settings are provided in the \textcolor{black}{supplemantary materials}.

\noindent\textbf{Finetuning Model Settings} Finetuning takes place by feeding the multispectral image tensor $\mathbf{x}$ as well as the sensor parameters $\{\boldsymbol{\lambda}_c, \sigma_c\}$ (if parameter encoding module is active) through the encoder network.
We denote the obtained features with index $i \in (0,\dots,N)$ ($N=C \times HW/P^2$) after layer $j$ as $\mathbf{f}_i^j = \encoder (\mathbf{x},\{\boldsymbol{\lambda}_c, \sigma_c\})$.
We take the set of features $\{ \mathbf{f}_i^3, \mathbf{f}_i^6, \mathbf{f}_i^9, \mathbf{f}_i^{12} \}$ and project all tokens originating from one image position into the latent dimension $D_{emb}$ via a linear layer, resulting in $HW/P^2$ projected features.
From here we follow \cite{ranftl2021vision} and use the projected features as inputs to a convolutional decoding network $\downstream$ to predict the segmentation mask. 
All weights for the parameter encoding module as well as the first $8$ transformer layers will be frozen for the finetuning step. 

Finetuning takes place over $150$ epochs with $5$ warmup-epochs, a cosign decay learning rate scheduler (initial learning rate of $1 \times 10^{-3}$), an \textit{AdamW optimizer}, a weight decay of 0.05 and processed with a batch size of $128$. Since the class distribution of the remaining eight classes from \textit{ESA-World Cover V200} over our validation area is highly imbalanced we apply batch sampling to approach an equal distribution. 
For evaluation purposes, we stick to an input sequence length of four channels which matches the number of channels during pre-training.

\section{Experiments \& Results}
\label{sec:results}

{
\setlength{\tabcolsep}{4pt}
\begin{table*}[t!]
    \centering
    \caption{Comparison of various pre-trained encoders for zero-shot evaluation. Following fine-tuning on S2\textsubscript{1237} to predict the landcover labels, the models are assessed in a zero-shot manner with respect to different sensors. All numbers represent micro IoU-scores. The checkmark indicates if the sensor parameter encoding and augmentation module were active (two checkmarks for using the sensor parameter encoding also before the decoding step).}
    \begin{tabular}{lccccccc}
    \toprule
    \multicolumn{4}{c}{} & \multicolumn{4}{c}{\textbf{Micro IoU}} \\
    \cmidrule(lr){5-8}
    \multicolumn{4}{c}{\textbf{Pre-Training}} & \textbf{Val. Set} & \multicolumn{3}{c}{\textbf{Zero Shot}} \\
    \cmidrule(lr){1-4} \cmidrule(lr){5-5} \cmidrule(lr){6-8}
    \footnotesize Backbone & \footnotesize Pretrained & \footnotesize\makecell{\footnotesize Sensor \\ Parameter \\ Encoding} & \footnotesize\makecell{ SSA \\ Module}  & S2\textsubscript{1237} & S2\textsubscript{1234} & SD\textsubscript{2468} & SD\textsubscript{1357}\\
    \cmidrule(lr){1-4} \cmidrule(lr){5-5} \cmidrule(lr){6-8}
    UNet &  & & & 0.69 & 0.39 & 0.47 & 0.21\\
    BaseMAE & & & & 0.69 & 0.40 & 0.41 & 0.17 \\
    \arrayrulecolor{black!30}\cmidrule(lr){1-4} \cmidrule(lr){5-5} \cmidrule(lr){6-8}
    BaseMAE & \checkmark &  & & 0.71 & 0.36 & 0.60 & 0.35 \\
    SenPa-MAE & \checkmark & \checkmark & & 0.68 & 0.67 & 0.58 & 0.41 \\
    SenPa-MAE & \checkmark & \checkmark & \checkmark & 0.70 & 0.68 & 0.64 & 0.46 \\
    SenPa-MAE & \checkmark & \checkmark \checkmark & & 0.71 & \textbf{0.69} & 0.63 & \textbf{0.52} \\
    SenPa-MAE & \checkmark & \checkmark \checkmark & \checkmark & 0.71 & \textbf{0.69} & \textbf{0.65} & \textbf{0.52} \\
    \arrayrulecolor{black}\bottomrule
    \end{tabular}
    \label{tab:S2Performance}
\end{table*}
}

{
\setlength{\tabcolsep}{4pt}
\begin{table*}[t!]
    \centering
    \caption{Comparison of various pre-trained encoders for training on inputs from varying sensors. Input data for finituning is given by random channels from either S2 or SD sensor. Corresponding evaluation micro IoU-scores on S2\textsubscript{1237}, SD\textsubscript{2468} and SD\textsubscript{1357} is given. The checkmark indicates if the sensor parameter encoding and augmentation module were active (two checkmarks for using the sensor parameter encoding also before the decoding step).}
    \begin{tabular}{lccccccc}
    \toprule
    \multicolumn{4}{c}{\textbf{Pre-Training}} & \multicolumn{4}{c}{\textbf{\makecell{Micro IoU}}} \\
    \cmidrule(lr){1-4} \cmidrule(lr){5-8}
    \footnotesize Backbone & \footnotesize Pretrained & \footnotesize\makecell{\footnotesize Sensor \\ Parameter \\ Encoding} & \footnotesize\makecell{ SSA \\ Module}  & S\textsubscript{rand} & S2\textsubscript{1237} & SD\textsubscript{2468} & SD\textsubscript{1357} \\ 
    \cmidrule(lr){1-4} \cmidrule(lr){5-8}
    UNet &  & &  & 0.69 & 0.70 & 0.72 & 0.70\\
    \arrayrulecolor{black!30}\cmidrule(lr){1-4} \cmidrule(lr){5-8}
    BaseMAE & \checkmark &  &  & 0.71 & 0.71 & 0.73 & 0.72 \\
    SenPa-MAE & \checkmark & \checkmark &  & 0.70 & 0.70 & 0.73 & 0.70 \\
    SenPa-MAE & \checkmark & \checkmark & \checkmark & 0.71 & 0.70 & 0.73 & 0.72 \\
    SenPa-MAE & \checkmark & \checkmark \checkmark &  & \textbf{0.72} & \textbf{0.72} & \textbf{0.74} & \textbf{0.73} \\
    SenPa-MAE & \checkmark & \checkmark \checkmark & \checkmark & 0.71 & 0.71 & 0.73 & 0.72 \\
    \arrayrulecolor{black}\bottomrule
    \end{tabular}
    \label{tab:RandomSensor}
\end{table*}
}

In this section, we test the capability of the pre-trained models to solve a landcover segmentation task in a multisensor arrangement. For the evaluation, we use data from the S2 and SD sensors and further simulate different synthetic sensor arrangements (which we refer to as different sensors) by splitting up the available data into different channel groups. 
We conduct two experiments where (a) the model is fine-tuned on one fixed input distribution (representing one fixed sensor) and the validation is conducted in a zero-shot manner for data from a different sensor and (b) a setting where one model is fine-tuned on data originating from multiple different sensors. 
All pre-trained models were exposed to data from all the available bands for each of the satellites shown in \cref{fig:responsefunctions}. For the downstream task experiments, we will refer to specific arrangements of a four-channel input for the sensors S2 and SD with the four indices as subscripts. Therefore, e.g. the 10m GSD bands of S2 get denoted as S2\textsubscript{1237} and the spectrally matching counterpart of the SD sensors as SD\textsubscript{2468} since the coastal blue channel is not present in the set of bands for the S2 satellite.

\noindent\textbf{Zero-Shot Experiments} We finetune the models to predict the landcover class given S2\textsubscript{1237}. We then conduct zero shot experiments with respect to a shift input distribution for a change in the infrared channel S2\textsubscript{1234} a change in spatial resolution SD\textsubscript{2468} as well as for the spectrally not aligning bands of the \textit{SuperDove} sensor SD\textsubscript{1357}. During inference, we feed the data as well as the corresponding sensor parameters (if parameter encoding is active) to the model. 
\Cref{tab:S2Performance} shows the results for the different encoders $\encoder$, by varying the type of sensor parameter encoding strategy and the usage of the SSA augmentation module.
As a baseline, we used a vanilla \textit{UNet} \cite{ronneberger2015u} as well as a non-pretrained version of the \textit{BaseMAE} architecture. 
\Cref{tab:S2Performance} reveals that, as expected, all setups achieve a satisfactory IoU score of $\approx 0.70$ when validated on data from the same distribution as the training set S2\textsubscript{1237}. However, conducting the validation in a zero-shot manner on the three different sensor arrangements (S2\textsubscript{1234}, SD\textsubscript{2468} and SD\textsubscript{1357}) drops significantly (up to 0.21) for the non-pre-trained baseline as well as for the \textit{BaseMAE} which does not encode sensor parameters during pretraining.
Models are pre-trained and tested with the parameter encoding module leading to more robust embeddings, specifically when applying the module twice before the encoding and decoding step. 
Moreover, in both instances (one or two active parameter encoding modules), implementing the SSA yields an additional improvement in performance, highlighting the effectiveness of the proposed data augmentation strategy.

\noindent\textbf{Finetuning on Multiple Sensors} We conduct the fine-tuning of the pre-trained models on varying sensors (with corresponding sensor parameters if the sensor parameter encoding module is active). We refer to S\textsubscript{random} to an input signal consisting of four randomly chosen channels from either S2 or SD (equal probably). This represents the most difficult task due to a random order of channels with varying geometric and spectral properties. We also present a slightly simpler setup with only two varying input sensors in the \textcolor{black}{supplemantary materials}.
\Cref{tab:RandomSensor} shows the validation IoU score for validating on S\textsubscript{random} as well as three fixed input data arrangements (S2\textsubscript{1234}, SD\textsubscript{2468} and SD\textsubscript{1357}). Similar to the findings in \cref{tab:S2Performance} pre-trained models with active sensor parameter encodings lead to the most robust embeddings with respect to predicting the downstream task. Graphical results for the segmentation task for all models can be found in the \textcolor{black}{supplemantary materials}.

\section{Conclusion and Outlook}
\label{sec:outlook}

This work introduces \textit{SenPa-MAE}, a transformer-based architecture that incorporates the sensor parameter of a multispectral satellite image into the image embeddings. 
To generate sensor-independent embeddings, pre-training is conducted over images from multiple sensors while taking the spectral response function as well as the ground sampling distance of the sensor into account.
We show that \textit{SenPa-MAE} produces more generative image embeddings via zero-shot evaluation and fine-tuning with subsequent testing on images from varying sensors. 

While our findings for the task of landcover mapping indicate that the incorporation of the sensor parameters yields beneficial results in the two conducted experiments, we anticipate even greater gains for tasks where spectral characteristics are more critical. Specifically for tasks where geometric features are less descriptive (like agriculture, water quality or soil parameter applications), the incorporation of the spectral response function is of higher importance. However, the empirical proof is beyond the scope of this study.

Further, this work is conducted on \textit{Sentinel-2}, \textit{Planet-Supedove} and \textit{Landsat} imagery, but can easily be extended to an arbitrary set of sensors due to its versatile encoding strategy. Applying this method to an even more diverse set of sensors, such as data from panchromatic or hyperspectral satellites, could yield even greater benefits.

\bibliographystyle{splncs04}
\bibliography{mybibliography}

\end{document}



\def\SubNumber{25}

\def\GCPRTrack{Special Track: Photogrammetry and remote sensing}

\title{Supplemantary Materials: SenPa-MAE: Sensor Parameter Aware Masked Autoencoder for Multi-Satellite Self-Supervised Pretraining}

\ifreview
\else
	\titlerunning{SenPa-MAE: Sensor Parameter Aware Masked Autoencoder}

	\author{Jonathan Prexl\orcidID{0009-0006-2560-5817} \and Michael Schmitt\orcidID{0000-0002-0575-2362}}
	
	\authorrunning{J. Prexl and M. Schmitt}
	
	\institute{Department of Aerospace Engineering, University of the Bundeswehr Munich, Munich, Germany
    \email{\{jonathan.prexl, michael.schmitt\}@unibw.de}}
\fi

\maketitle              

\FloatBarrier
\section{Geographical Distribution of the Dataset}
\begin{figure}
    \centering
    \includegraphics[width=.7\textwidth]{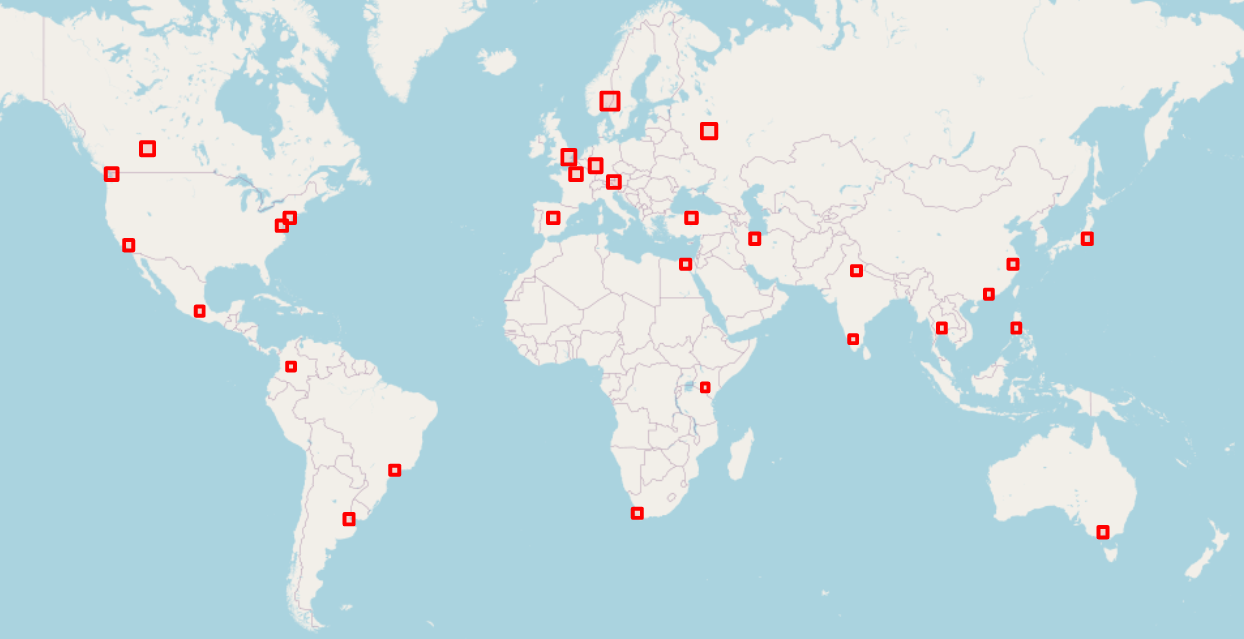}
    \caption{Distribution of the locations for the pre-training dataset.}
    \label{fig:location1}
\end{figure}

\begin{figure}
    \centering
    \includegraphics[width=.7\textwidth]{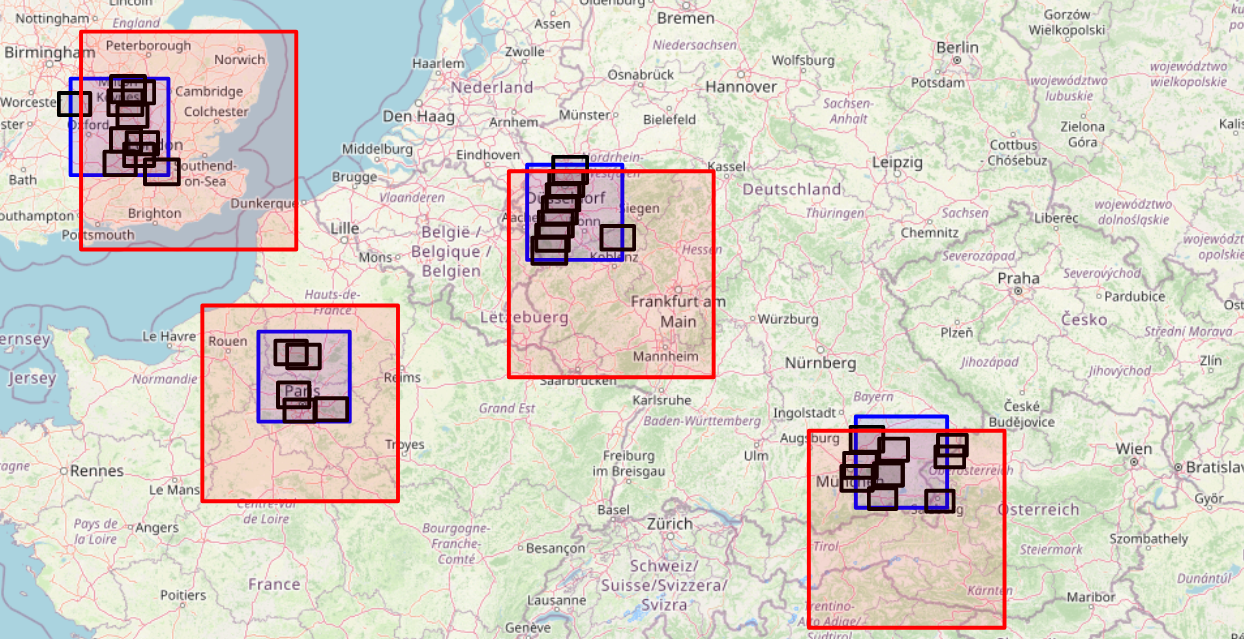}
    \caption{Zoomed view for the distribution of the locations for the pre-training dataset.}
    \label{fig:location2}
\end{figure}

\begin{figure}
    \centering
    \includegraphics[width=.7\textwidth]{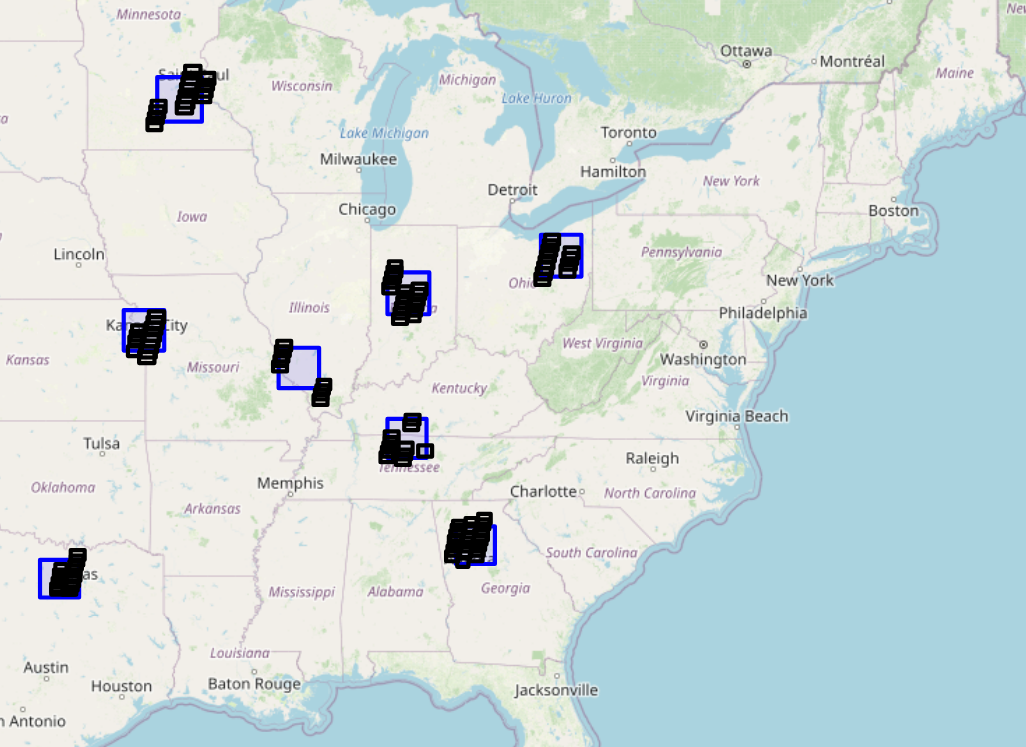}
    \caption{Distribution of the region of interest for the downstream task dataset where pre-trained models get fine-tuned to predict the land use according to ESA WorldCover.}
    \label{fig:location3}
\end{figure}

As described in the main body we sample images along 29 globally distributed locations for pre-training and eight locations in the US for the downstream task experiments. \Cref{fig:location1,fig:location2,fig:location3} display the geografical locations of the data. The red bounding boxes indicate the footprint of the \textit{Landsat} data, whereas blue indicates the footprints of the \textit{Sentinel-2} data and black polygons of the \textit{SuperDove} data, respectively. 

\FloatBarrier
\section{Addional Reconstruction Results}
\begin{figure}
    \centering
    \includegraphics[width=.5\textwidth]{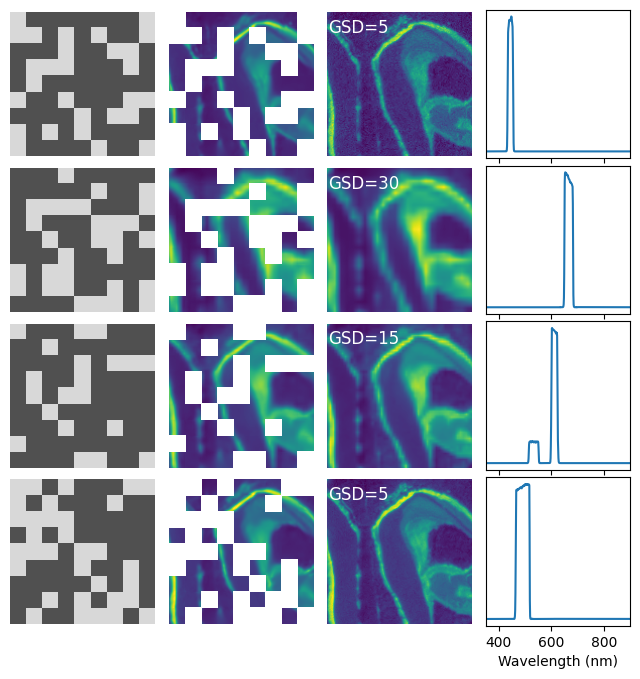}
    \caption{Visual example for the reconstruction task during MAE pre-training.}
    \label{fig:r1}
\end{figure}

\begin{figure}
    \centering
    \includegraphics[width=.5\textwidth]{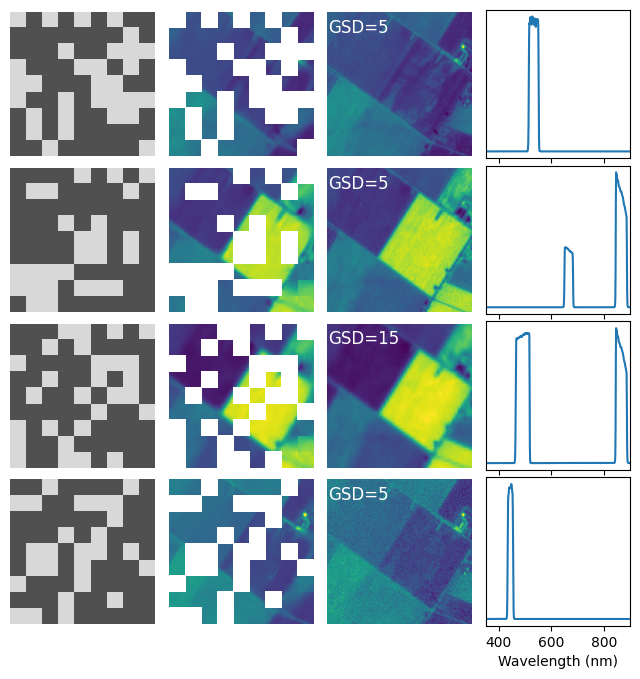}
    \caption{Visual example for the reconstruction task during MAE pre-training.}
    \label{fig:r2}
\end{figure}

\begin{figure}
    \centering
    \includegraphics[width=.5\textwidth]{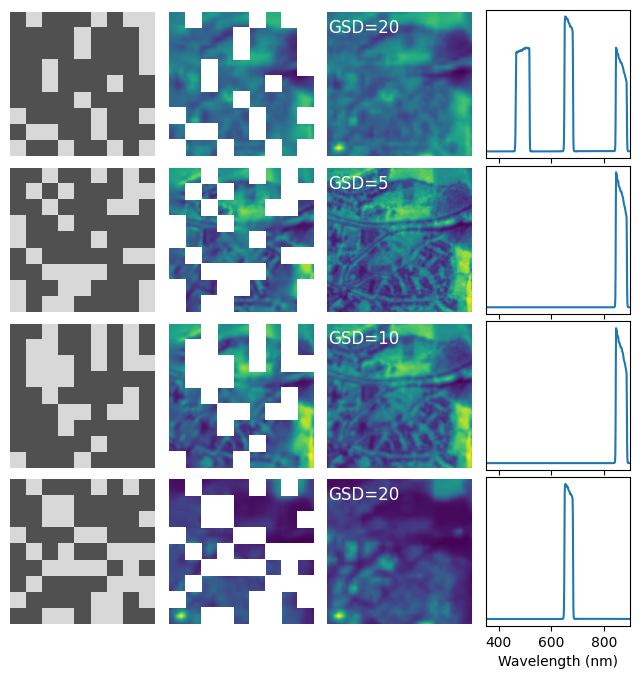}
    \caption{Visual example for the reconstruction task during MAE pre-training.}
    \label{fig:r3}
\end{figure}

\begin{figure}
    \centering
    \includegraphics[width=.5\textwidth]{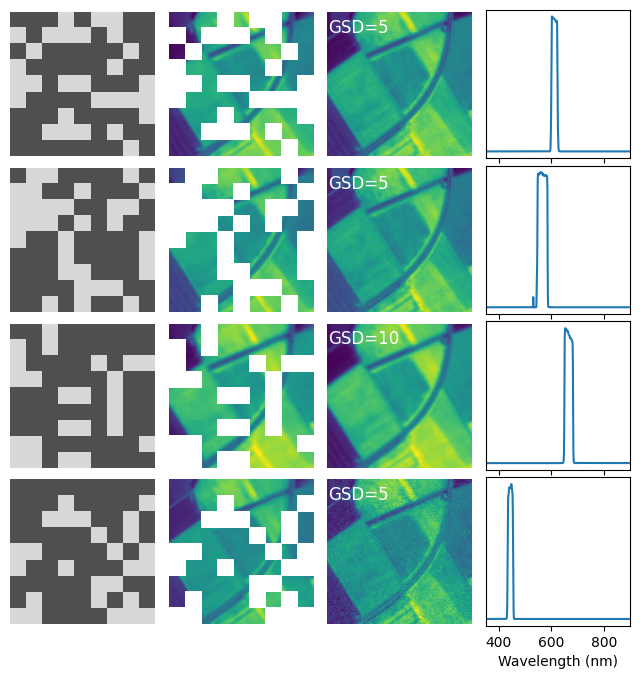}
    \caption{Visual example for the reconstruction task during MAE pre-training.}
    \label{fig:r4}
\end{figure}

\begin{figure}
    \centering
    \includegraphics[width=.5\textwidth]{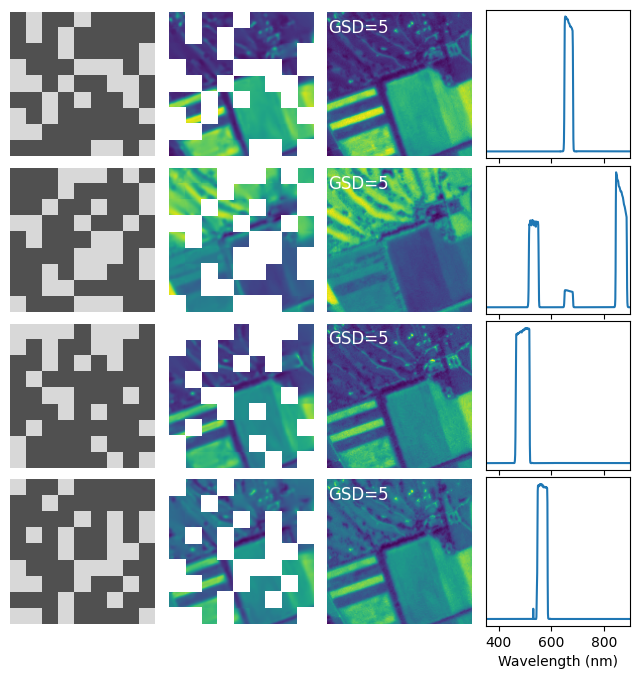}
    \caption{Visual example for the reconstruction task during MAE pre-training.}
    \label{fig:r5}
\end{figure}

\begin{figure}
    \centering
    \includegraphics[width=.5\textwidth]{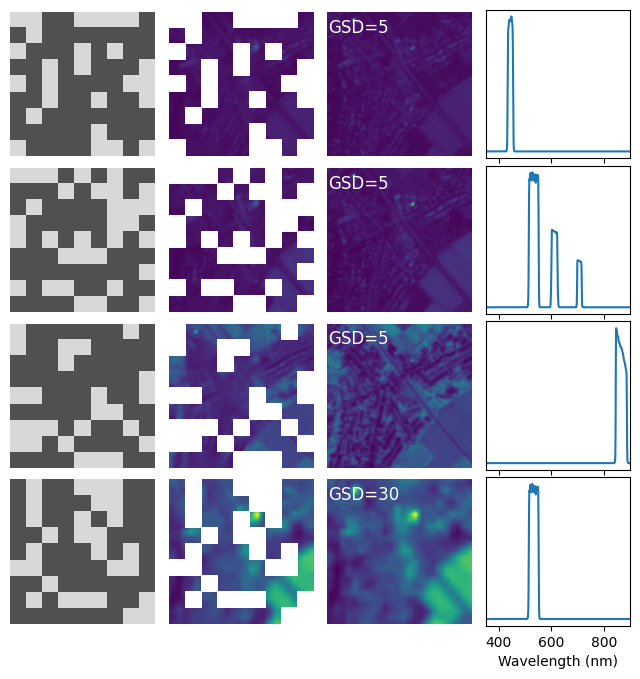}
    \caption{Visual example for the reconstruction task during MAE pre-training.}
    \label{fig:r6}
\end{figure}

In line with Figure 3 of the main manuscript, we show further additional reconstruction examples in \cref{fig:r1,fig:r2,fig:r3,fig:r4,fig:r5,fig:r6}. All images show from left to right: Channel-specific mask, reconstructed signal for the non-masked areas, ground truth image with the GSD of the channel, as well as the corresponding spectral response function.
The images correspond to the pertaining of \textit{SenPa-MAE} with dual parameter embedding module and active spectral superposition augmentation.


\FloatBarrier
\section{Finetuning on two alternating sensors}

Building on the finetuning experiment outlined in the main manuscript, where model inputs were composed of four randomly selected channels from either the S2 or SD sensors, here finetuning results using a fixed two-sensor configuration are presented. In this setup, we alternated between S2\textsubscript{1237} and SD\textsubscript{1357} as inputs to finetune the model for predicting land cover classes according to the ESA WorldCover. As shown in \cref{tab:DualSensor}, the model's performance follows a similar trend to the earlier experiment, though with slightly reduced absolute differences, likely because the fixed two-sensor input is easier for the model to adapt to.

{
\setlength{\tabcolsep}{4pt}
\begin{table*}[h!]
    \centering
    \caption{Comparison of various pre-trained encoders for training on inputs from varying sensors. Input data for finituning is given by alternatingly S2\textsubscript{1237} and SD\textsubscript{1357}. Corresponding evaluation micro IoU-scores on the two sensors are given. The checkmark indicates if the sensor parameter encoding and augmentation module were active (two checkmarks for using the sensor parameter encoding also before the decoding step).}
    \begin{tabular}{lccccc}
    \toprule
    \multicolumn{4}{c}{\textbf{Pre-Training}} & \multicolumn{2}{c}{\textbf{\makecell{Micro IoU}}} \\
    \cmidrule(lr){1-4} \cmidrule(lr){5-6}
    \footnotesize Backbone & \footnotesize Pretrained & \footnotesize\makecell{\footnotesize Sensor \\ Parameter \\ Encoding} & \footnotesize\makecell{ SSA \\ Module}  & S2\textsubscript{1237} & SD\textsubscript{1357} \\ 
    \cmidrule(lr){1-4} \cmidrule(lr){5-6}
    UNet &  & &  & 0.71 & 0.70 \\
    \arrayrulecolor{black!30}\cmidrule(lr){1-4} \cmidrule(lr){5-6}
    BaseMAE & \checkmark &  &  & 0.73 & 0.71  \\
    SenPa-MAE & \checkmark & \checkmark &  & 0.71 & 0.69  \\
    SenPa-MAE & \checkmark & \checkmark & \checkmark & 0.71 & 0.70  \\
    SenPa-MAE & \checkmark & \checkmark \checkmark &  & \textbf{0.73} & \textbf{0.71}  \\
    SenPa-MAE & \checkmark & \checkmark \checkmark & \checkmark & 0.72 & 0.70  \\
    \arrayrulecolor{black}\bottomrule
    \end{tabular}
    \label{tab:DualSensor}
\end{table*}
}

\FloatBarrier
\section{Addional Landcover Prediction Results}
\begin{figure}
    \centering
    \begin{subfigure}{0.49\textwidth}
        \includegraphics[width=\textwidth]{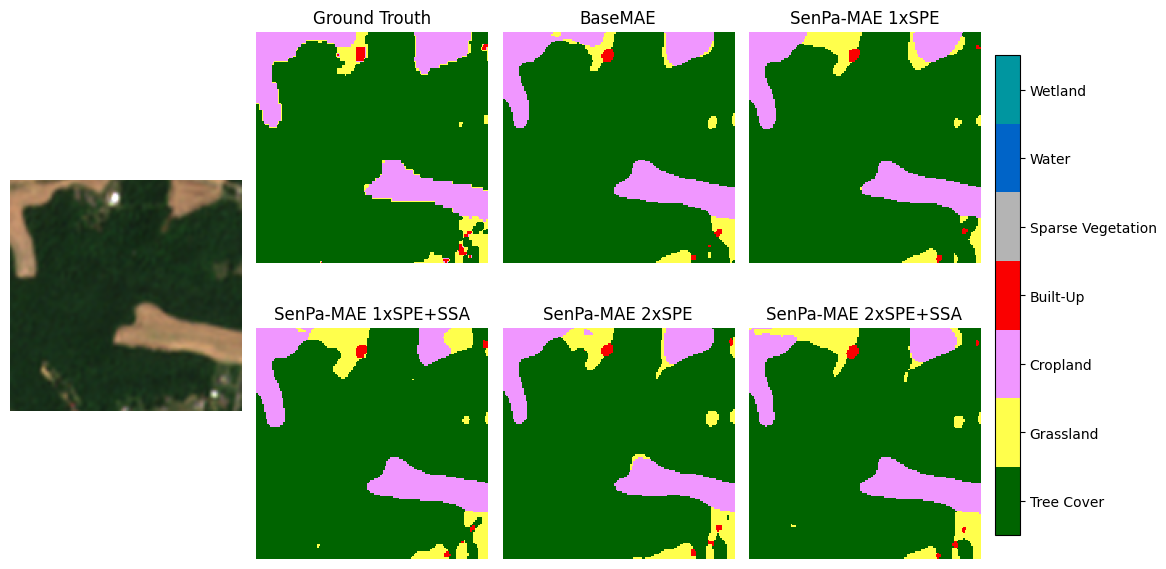}
        \caption{}
        \label{fig:first}
    \end{subfigure}
    \hfill
    \begin{subfigure}{0.49\textwidth}
        \includegraphics[width=\textwidth]{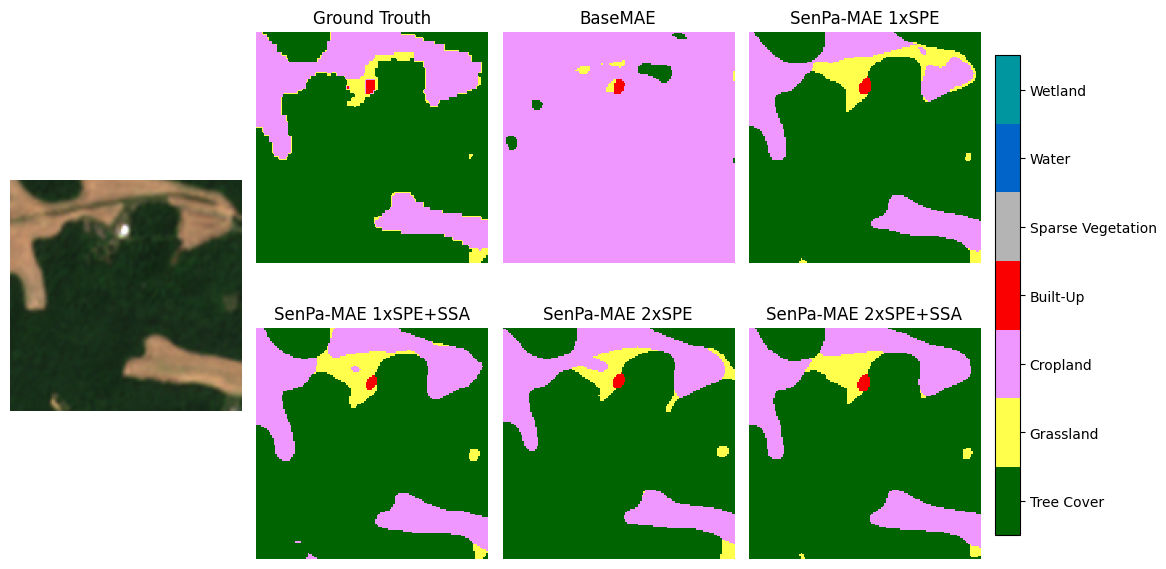}
        \caption{}
        \label{fig:second}
    \end{subfigure}
    \caption{Visual results for changing the channels of the input images during evaluation. Models trained on S2\textsubscript{1237} with subsequent evaluation on a.) S2\textsubscript{1237} and b.) S2\textsubscript{1234}.}
    \label{fig:ve1}
\end{figure}

\begin{figure}
    \centering
    \begin{subfigure}{0.49\textwidth}
        \includegraphics[width=\textwidth]{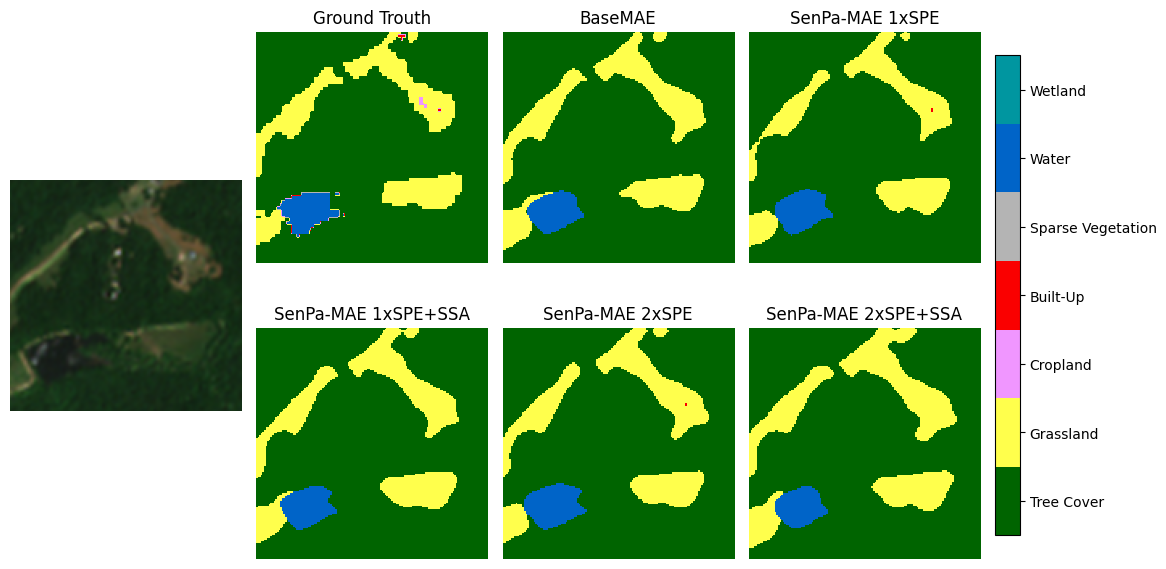}
        \caption{}
        \label{fig:first}
    \end{subfigure}
    \hfill
    \begin{subfigure}{0.49\textwidth}
        \includegraphics[width=\textwidth]{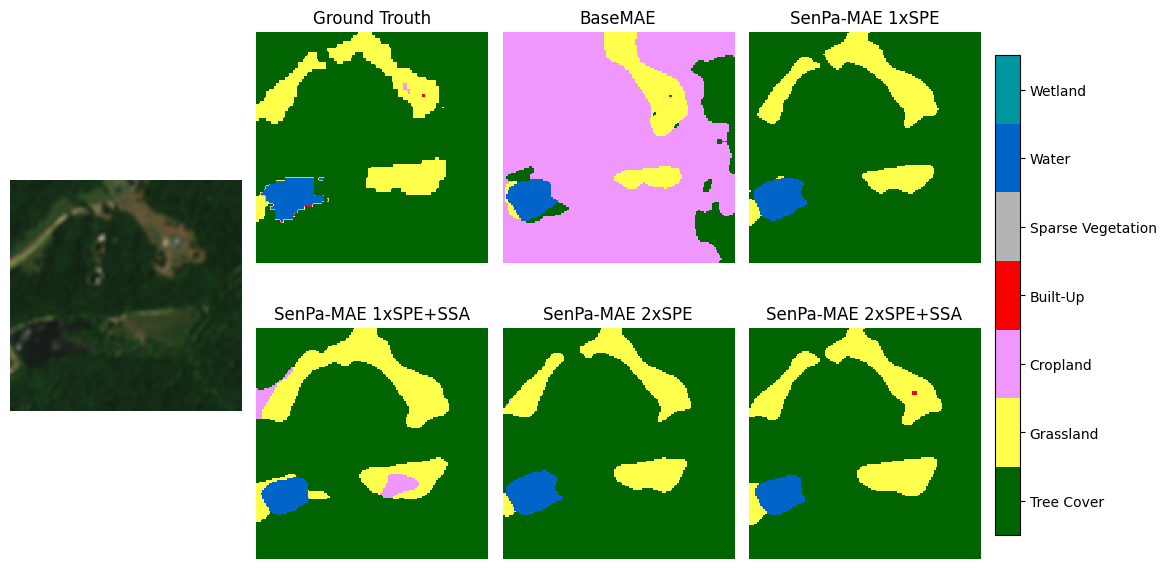}
        \caption{}
        \label{fig:second}
    \end{subfigure}
    \caption{Visual results for changing the channels of the input images during evaluation. Models trained on S2\textsubscript{1237} with subsequent evaluation on a.) S2\textsubscript{1237} and b.) S2\textsubscript{1234}.}
    \label{fig:ve2}
\end{figure}

\begin{figure}
    \centering
    \begin{subfigure}{0.49\textwidth}
        \includegraphics[width=\textwidth]{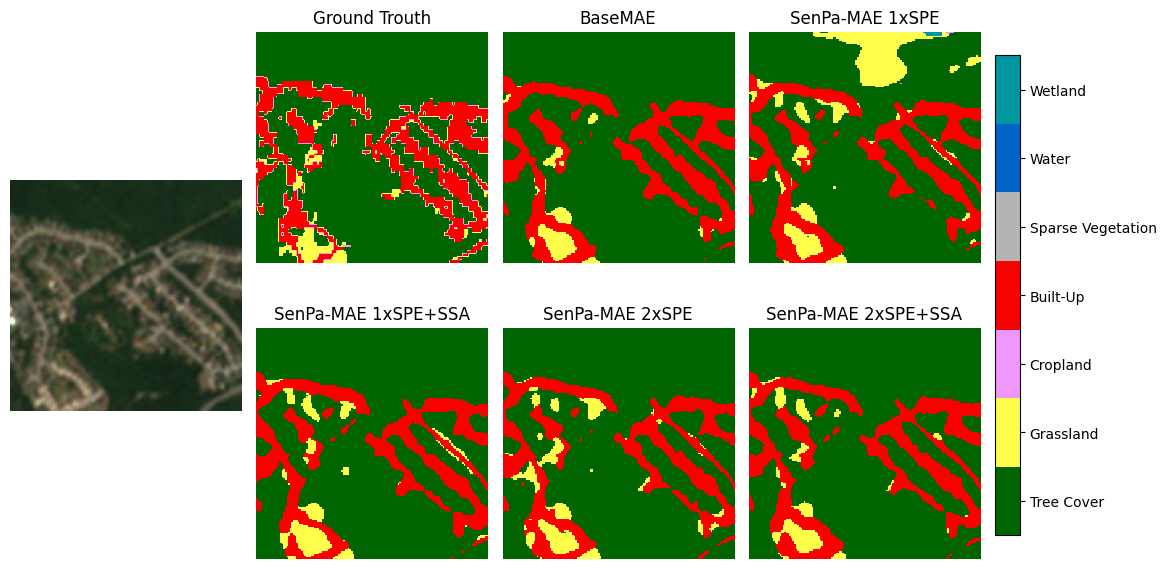}
        \caption{}
        \label{fig:first}
    \end{subfigure}
    \hfill
    \begin{subfigure}{0.49\textwidth}
        \includegraphics[width=\textwidth]{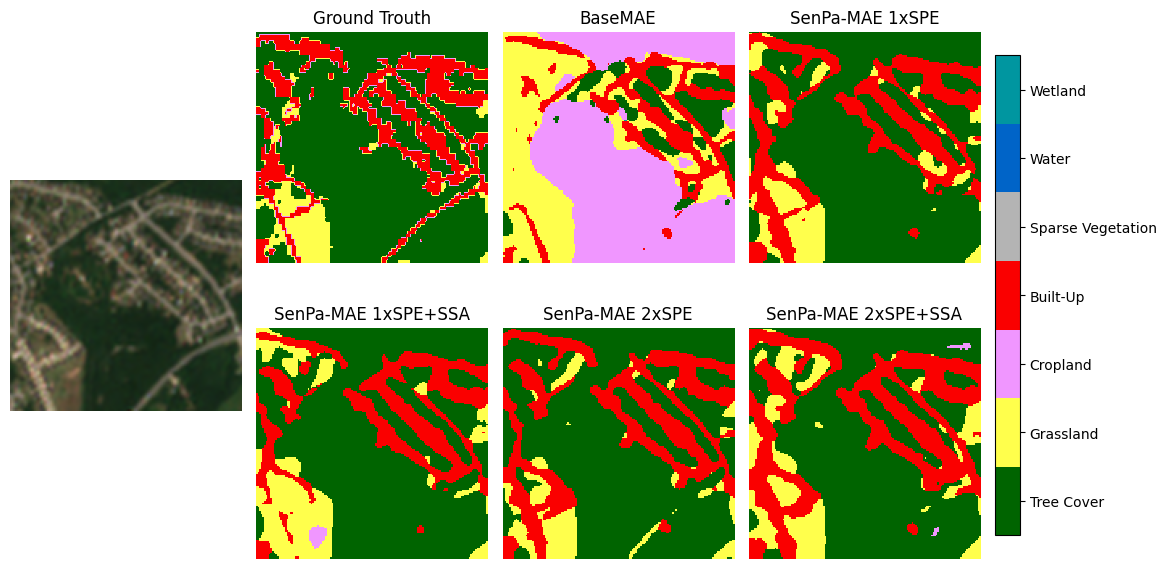}
        \caption{}
        \label{fig:second}
    \end{subfigure}
    \caption{Visual results for changing the channels of the input images during evaluation. Models trained on S2\textsubscript{1237} with subsequent evaluation on a.) S2\textsubscript{1237} and b.) S2\textsubscript{1234}.}
    \label{fig:ve3}
\end{figure}

\begin{figure}
    \centering
    \begin{subfigure}{0.49\textwidth}
        \includegraphics[width=\textwidth]{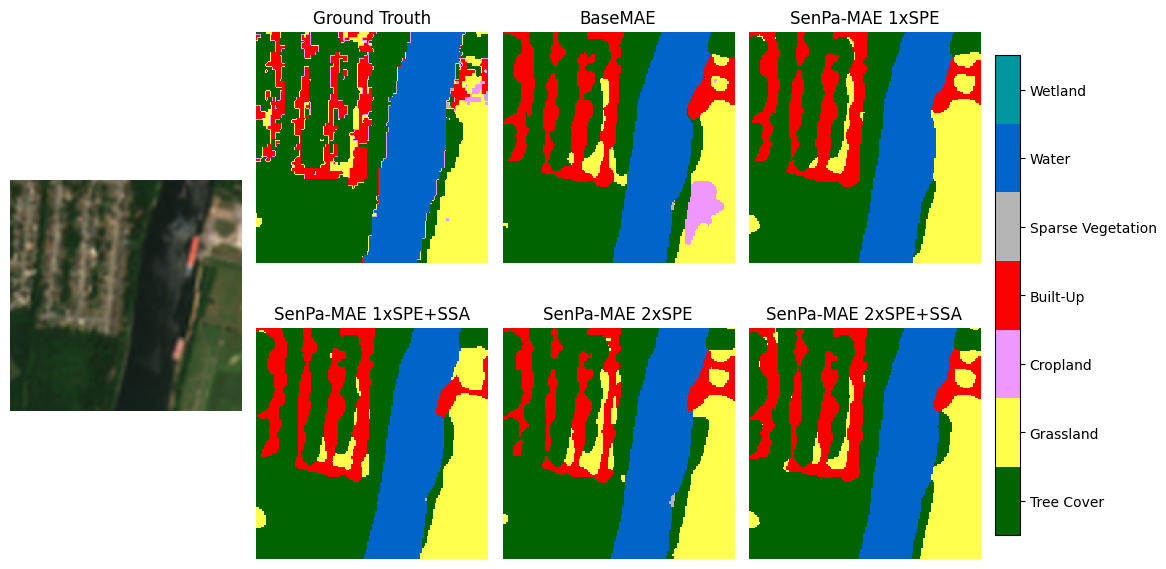}
        \caption{}
        \label{fig:first}
    \end{subfigure}
    \hfill
    \begin{subfigure}{0.49\textwidth}
        \includegraphics[width=\textwidth]{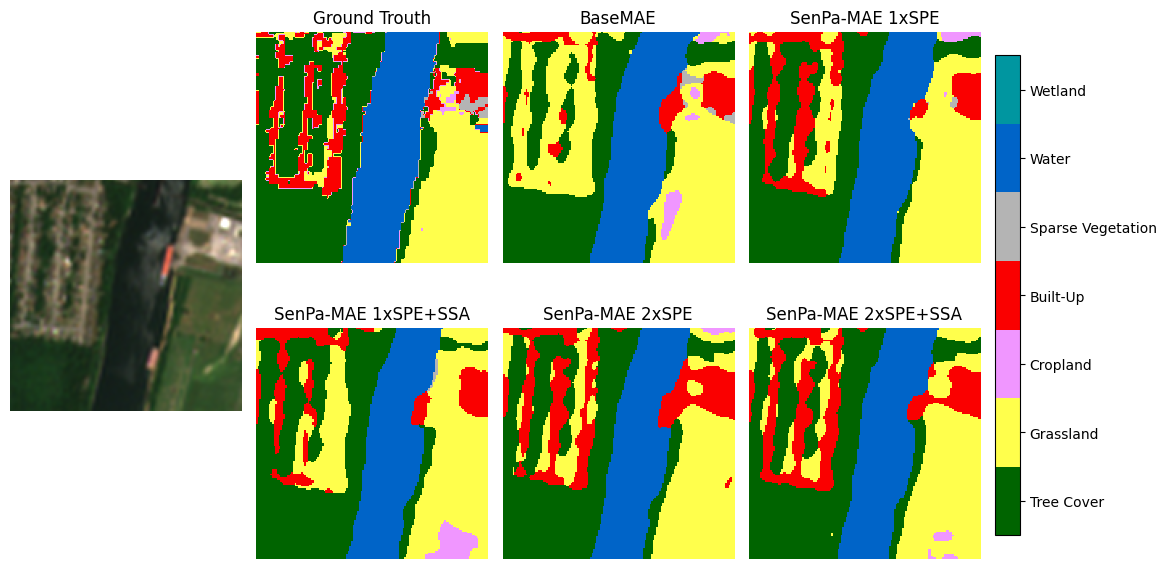}
        \caption{}
        \label{fig:second}
    \end{subfigure}
    \caption{Visual results for changing the channels of the input images during evaluation. Models trained on S2\textsubscript{1237} with subsequent evaluation on a.) S2\textsubscript{1237} and b.) S2\textsubscript{1234}.}
    \label{fig:ve4}
\end{figure}

Following, visual examples of the performance of the landcover segmentation problem, used as a downstream task evaluation for our proposed method are shown.
For each of the examples in \cref{fig:ve1,fig:ve2,fig:ve3,fig:ve4} we show the predicted landcover mask for with and without the distribution shift of the zero-shot evaluation experiments. 
\textit{SSA} denotes the active augmentation module and \textit{1 $\times$ SPE} or \textit{2 $\times$ SPE} denotes the application of the sensor parameter encoding step (only before the encoder step or twice for the encoder as well as the decoder step).

Whereas all models perform similarly when evaluated with images form training data distribution S2\textsubscript{1237} (left column in \cref{fig:ve1,fig:ve2,fig:ve3,fig:ve4}) models trained with the parameter encoding module show a more robust behaviour when evaluated with S2\textsubscript{1234} (right column in \cref{fig:ve1,fig:ve2,fig:ve3,fig:ve4}). 

